\documentclass[runningheads]{llncs}

\usepackage{iciap}

\usepackage{iciapabbrv}

\usepackage{graphicx}
\usepackage{booktabs}

\usepackage[accsupp]{axessibility}  

\usepackage{hyperref}

\usepackage{orcidlink}

\begin{document}

\title{An Investigation of Ear-EEG Signals for a Novel Biometric Authentication System} 

\titlerunning{Abbreviated paper title}

\author{Danilo Avola\inst{1}\orcidlink{0000-0001-9437-6217} \and
Giancarlo Crocetti\inst{2}\orcidlink{0000-0001-9437-6217} \and
Gian Luca Foresti\inst{3}\orcidlink{0000-0002-8425-6892} \and \\
Daniele Pannone\inst{1}\orcidlink{0000-0001-6446-6473} \and 
Claudio Piciarelli\inst{3}\orcidlink{0000-0001-5305-1520} \and 
Amedeo Ranaldi\inst{1} \and \\ Davide Vittucci\inst{1} }

\authorrunning{Avola et al.}

\institute{Department of Computer Science, Sapienza University, Italy \\
\email{\{avola, pannone, ranaldi, vittucci\}@di.uniroma1.it} \and
Department of Computer Science and Mathematics, \\
The L. H. and W. L. Collins College of Professional Studies, \\ 
St. John's University, USA.\\
\email{crocettg@stjohns.edu}\\
\and
Department of Mathematics and Computer Science, University of Udine, Italy
\email{\{gianluca.foresti,claudio.piciarelli\}@uniud.it}}

\maketitle

\begin{abstract}
This work explores the feasibility of biometric authentication using EEG signals acquired through in-ear devices, commonly referred to as ear-EEG. Traditional EEG-based biometric systems, while secure, often suffer from low usability due to cumbersome scalp-based electrode setups. In this study, we propose a novel and practical framework leveraging ear-EEG signals as a user-friendly alternative for everyday biometric authentication. The system extracts an original combination of temporal and spectral features from ear-EEG signals and feeds them into a fully connected deep neural network for subject identification. Experimental results on the only currently available ear-EEG dataset suitable for different purposes, including biometric authentication, demonstrate promising performance, with an average accuracy of 82\% in a subject identification scenario. These findings confirm the potential of ear-EEG as a viable and deployable direction for next-generation real-world biometric systems.
\end{abstract}

\section{Introduction}
\label{sec:intro}
Biometric authentication has become a critical component of modern security systems, providing reliable and user-friendly mechanisms for identity verification in various application domains. Traditional approaches such as fingerprint scanning~\cite{9581287}, facial recognition~\cite{10.1145/954339.954342}, and iris detection~\cite{10.1145/3651306} have gained widespread adoption; however, they are not without limitations. Many of these systems are vulnerable to spoofing, degradation over time, or environmental interference. To overcome these challenges, recent research has focused on bio-signals that are inherently more secure and less prone to manipulation, such as ElectroEncephaloGraphy (EEG) signals~\cite{ACHARYA2013147,AVL_01}. EEG-based authentication relies on the uniqueness and complexity of brain signals, which are extremely difficult to replicate or forge~\cite{Wu2018}. Unlike biometric traits that can be externally captured or copied, EEG patterns reflect internal neurological responses that vary subtly across individuals~\cite{7002950}. These signals also exhibit resistance to stress, fatigue, or environmental noise, making them a promising modality for secure user identification and verification~\cite{Custom_1}. Moreover, EEG offers the opportunity for continuous authentication and liveness detection, overcoming limitations of static biometrics like fingerprints or facial features. Nonetheless, one major barrier to real-world adoption is the impracticality of conventional scalp-EEG setups, which involve multiple electrodes, conductive gels, and often uncomfortable headgear. To address these usability issues, the concept of ear-EEG has emerged~\cite{10.3389/fnins.2024.1441897}. Ear-EEG uses electrodes embedded in in-ear devices to capture brain signals from within the ear canal. This setup is significantly more user-friendly and unobtrusive, aligning with the form factor of commonly used earbuds or hearing aids. While ear-EEG has shown promise in various fields including sleep monitoring, drowsiness detection, and brain-computer interfaces, its application to biometric authentication remains largely unexplored. The possibility of recording high-quality EEG signals from the ear canal with dry electrodes opens new perspectives for practical and minimally invasive biometric systems.

In this study, we propose a novel biometric authentication framework based on ear-EEG signals. Drawing on selected contributions from the literature on EEG feature extraction and sequence analysis, including works focused on spectral descriptors~\cite{8067531}, temporal dynamics~\cite{9967652}, and the semantic complexity and interpretability of EEG sequences~\cite{AVL_01,AVL_02}, our approach extracts a unique combination of four feature types: Power Spectral Density (PSD), Autoregressive Coefficients (AR), Hjorth parameters, and Spectral Entropy, laying the groundwork for a compact yet expressive representation of brain activity. These features are processed by a fully connected Deep Neural Network (DNN), designed specifically to model non-linear relationships among descriptors and to classify subjects effectively. In previous studies involving ear-EEG for various classification tasks, traditional machine learning models such as Support Vector Machines (SVMs) and Linear Discriminant Analysis (LDA) have commonly been adopted~\cite{8067531,s24186004}. In contrast, our approach introduces a deep architecture within a biometric authentication framework, thus offering greater flexibility and enhanced learning capacity, particularly in effectively modeling inter-subject variability.

Given the scarcity of publicly available ear-EEG datasets for authentication purposes, we leverage the only current dataset that can be adapted to this task: a collection originally designed for motor task classification~\cite{Wu_2020}. While not natively biometric, this dataset includes high-quality ear-EEG recordings from multiple subjects, making it suitable for repurposing in identity recognition studies. Moreover, in contrast to comparable works that rely on private datasets, thus limiting reproducibility and access, our use of a public resource enhances the transparency and repeatability of the research. Although the original acquisitions were collected for motor-related tasks, the ear-EEG signals captured reflect subject-specific brain dynamics. Supporting this, prior work on motor imagery using conventional EEG setups~\cite{Avola202483940} has shown that such signals can be robustly and consistently classified across individuals, highlighting their potential suitability for biometric applications. To further improve the reliability and generalizability of our system, we also apply a structured pre-processing and augmentation pipeline, including Gaussian noise injection, temporal shifting, MixUp augmentation, random oversampling, and class weighting. Concluding, the main contributions of this work can be summarized as follows:
\begin{itemize}
    \item We propose a new feature set combining PSD, AR, Hjorth parameters, and Spectral Entropy to better represent EEG signals for identity recognition.
    \item We design and implement a deep classifier specifically tailored for subject identification using ear-EEG signals, in contrast with prior approaches in this domain that typically rely on shallow models such as SVM or LDA.
    \item We implement a structured data pre-processing and augmentation strategy to improve robustness and mitigate class imbalance.
\end{itemize}

These contributions demonstrate the viability of ear-EEG for biometric applications and provide a strong foundation for future developments in real-world, user-friendly authentication systems.

\section{Related Work}
EEG has emerged as a promising biometric modality due to its ability to capture intrinsic and non-replicable neural patterns. Unlike conventional traits such as fingerprints or facial features, EEG signals reflect individual brain activity that is not externally observable, making them harder to spoof, and potentially suitable for continuous authentication in real-world scenarios~\cite{ACHARYA2013147,6748964}. In particular, EEG signals possess a high degree of inter-subject variability while maintaining intra-subject consistency under controlled conditions, two characteristics that are crucial for the design of biometric systems. Early studies, as summarized in~\cite{6748964}, have reported promising results in EEG-based biometric recognition, with various feature extraction techniques, including spectral, temporal, and statistical descriptors, thus demonstrating the potential of EEG signals for subject identification. These approaches were typically applied to resting-state EEG acquired through scalp electrodes and involved classical machine learning classifiers. Moving toward more advanced modeling strategies, recent works, as reviewed in~\cite{Shams_Hossain}, have highlighted the growing use of deep learning techniques in EEG-based biometric systems. Architectures such as Convolutional Neural Networks (CNNs) and Long Short-Term Memory (LSTM) networks have shown the potential to further improve recognition performance, particularly when large-scale EEG datasets are available. Despite these advancements, the practical deployment of EEG biometrics has been limited by the inconvenience of traditional scalp-based acquisition, which typically involves wet electrodes, time-consuming setup, and low wearability. In recent years, the search for more practical and user-friendly acquisition methods has led to the exploration of ear-EEG as a promising alternative. In this approach, electrodes are embedded into wearable earpiece-like devices to enable portable and unobtrusive signal acquisition.

A notable contribution within this emerging line of research is provided in~\cite{8067531}, which demonstrated the feasibility of using ear-EEG for person authentication by extracting temporal and spectral descriptors, and applying traditional classifiers such as SVM and LDA. Another significant example is~\cite{s24186004}, where ear-EEG signals were employed for motor task classification using standard machine learning techniques. However, many of these studies are based on non-public datasets, which limits reproducibility and comparative assessment across the field. Furthermore, deep neural architectures remain underexplored in the context of ear-EEG for biometric authentication. A recent doctoral work~\cite{HWIDI2023} represents an initial step in this direction, proposing a proof-of-concept in-ear EEG system and exploring the applicability of deep learning techniques such as CNNs, LSTM networks, Variational Autoencoders (VAEs), and MiniRocket. While these models are primarily applied to conventional EEG data within the study, the in-ear EEG component focuses on feature extraction and classification using a custom dataset collected over multiple sessions. The present work aims to address these limitations by applying a fully connected DNN to ear-EEG signals obtained from a publicly accessible dataset initially developed for motor imagery classification~\cite{Wu_2020}. In addition, a structured pipeline of feature engineering and data augmentation is introduced to improve model robustness and generalization. By relying on an open and publicly accessible dataset, adopting a deep learning-based classification framework, and introducing a comprehensive and well-structured feature extraction strategy that incorporates PSD, AR, Hjorth parameters, and Spectral Entropy, this study offers a reproducible and scalable contribution to the field of biometric authentication using ear-EEG.

\section{Proposed Method}
\label{sec:proposed_method}
\begin{figure}[t]
    \centering
    \includegraphics[width=\linewidth]{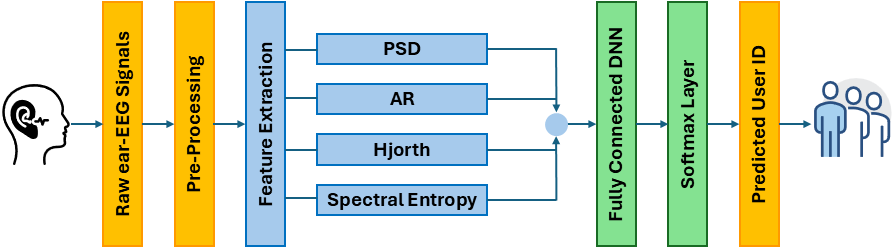}
    \caption{Overview of the proposed architecture. Raw ear-EEG signals undergo pre-processing and feature extraction using PSD, AR, Hjorth parameters, and Spectral Entropy. The features are concatenated and fed into a fully connected DNN that predicts user identity via softmax classification.}
    \label{Fig:01}
\end{figure}
Fig.~\ref{Fig:01} provides a high-level overview of the proposed biometric authentication framework based on ear-EEG signals. The system comprises three main components: signal pre-processing, feature extraction, and classification via a fully connected DNN; each will be discussed in detail in the following subsections.
\subsection{Pre-Processing}
\label{secsec:pre-processing}
The proposed system relies on the publicly available ear-EEG dataset introduced in~\cite{Wu_2020}, originally designed for motor task classification. The acquisitions were collected from eight in-ear electrodes, namely LF, LB, LOU, LOD, RF, RB, ROU, and ROD, placed bilaterally around the ear canal and sampled at 1,000~Hz. The spatial configuration of the ear-EEG electrodes follows the setup describe in~\cite{Wu_2020}, where their relative positions around the ear canal are illustrated in detail. A total of six subjects performed a variety of motor tasks during the acquisition sessions, resulting in a diverse set of EEG recordings that can be repurposed for biometric authentication. To prepare the signals for feature extraction, a pre-processing pipeline is applied. First, only the channels corresponding to the in-ear region are selected from the full montage. A bandpass filter in the range of 0.5-100~Hz is then used to isolate the frequency components relevant to typical EEG activity. Additionally, a 50~Hz notch filter is applied to suppress power line interference. The resulting cleaned signals serve as the basis for the subsequent analysis. Formally, let $x_c(t)$ denote the continuous EEG signal recorded from channel $c \in \{1, \dots, C\}$, where $C = 8$ in this case. The signal is segmented into overlapping windows of fixed duration $T$, each window corresponding to a discrete sequence $\mathbf{x}_c = [x_c(1), x_c(2), \dots, x_c(N)]$ of $N$ samples. These windowed segments are then used as input for feature extraction. 

Although the dataset includes recordings from only six subjects, this limitation is effectively mitigated by the intrinsic richness of EEG data, which are recorded as high-dimensional temporal sequences. Each subject contributes a large number of signal windows, allowing the model to learn fine-grained, subject-specific patterns over time. In addition, as will be detailed in the experimental section, an extensive and targeted data augmentation strategy has been employed to synthetically expand the dataset. The augmentation techniques have been carefully designed to preserve the temporal dynamics and spectral properties of the original signals, thus ensuring that the generated samples remain representative and suitable for training a robust biometric classification model.
\subsection{Feature Extraction}
\label{secsec:feature_extraction}
To characterize the biometric identity encoded in ear-EEG signals, we extract a structured set of features that capture both spectral and temporal information. From each pre-processed signal segment $\mathbf{x}_c = [x_c(1), x_c(2), \dots, x_c(N)]$ acquired from channel $c$, we compute four descriptors: Power Spectral Density (PSD), Autoregressive (AR) Coefficients, Hjorth Parameters, and Spectral Entropy.
\paragraph{Power Spectral Density (PSD):}
PSD is a widely used technique in EEG analysis as it reveals how the energy of the signal is distributed across frequency bands. This feature helps characterize individual-specific spectral patterns. We estimate the PSD using the Welch method~\cite{1161901}, which improves stability by reducing variance compared to the standard Fast Fourier Transform (FFT). Specifically, each EEG segment is divided into overlapping windows of 256 samples, and the average periodogram is computed. Given the sampling rate of 1,000~Hz, this window size provides a balanced trade-off between temporal resolution and spectral accuracy. Since most relevant biometric information resides in the lower frequency bands, only the first 20 spectral coefficients are retained, yielding a compact and informative representation. The PSD estimate for channel $c$ can be defined as follows:
\begin{equation}
\widehat{P}_c(f) = \frac{1}{K} \sum_{k=1}^{K} \left| \mathcal{F}\left(w_k[n] \cdot x_c^{(k)}[n] \right) \right|^2,
\end{equation}
where $\widehat{P}_c(f)$ is the estimated power spectral density for channel $c$ at frequency $f$, $K$ is the number of overlapping subsegments extracted from the original EEG window, $x_c^{(k)}[n]$ denotes the $k$-th subsegment of the signal from channel $c$, $w_k[n]$ is the windowing function applied to that subsegment, and $\mathcal{F}$ represents the discrete Discrete Fourier Transform (DFT). Each $\left| \cdot \right|^2$ term corresponds to the periodogram of a subsegment, and the average over $K$ segments improves the robustness of the spectral estimate by reducing its variance.
\paragraph{Autoregressive (AR) Coefficients:}
To capture the temporal dependencies in EEG signals, we compute AR coefficients, which model each sample as a linear combination of its past values. This allows the system to exploit the sequential structure of the data for identity recognition. We use the Yule-Walker method~\cite{1927RSPTA.226..267U} for coefficient estimation due to its computational efficiency and numerical stability. The AR model of order $p$ for channel $c$ is expressed as:
\begin{equation}
x_c(n) = -\sum_{i=1}^{p} a_i x_c(n-i) + \varepsilon(n),
\end{equation}
where $a_i$ are the AR coefficients and $\varepsilon(n)$ is the prediction error. The estimated vector $\mathbf{a}_c = [a_1, a_2, \dots, a_p]$ is included as a feature set.
\paragraph{Hjorth Parameters:}
The Hjorth descriptors provide a compact time-domain characterization of EEG signals through three components: Activity, Mobility, and Complexity. These parameters quantify amplitude variance, frequency content, and waveform irregularity, respectively. Let $\mathbf{x}_c$ be a signal segment from channel $c$, and let $\mathbf{x}_c'$, $\mathbf{x}_c''$ denote its first and second temporal derivatives. The Hjorth parameters are defined as:
{\small
\begin{equation}
\text{Activity} = \text{Var}(\mathbf{x}_c), \quad
\text{Mobility} = \sqrt{ \frac{\text{Var}(\mathbf{x}_c')}{\text{Var}(\mathbf{x}_c)} }, \quad
\text{Complexity} = \frac{\text{Mobility}(\mathbf{x}_c')}{\text{Mobility}(\mathbf{x}_c)}.
\end{equation}
}
\paragraph{Spectral Entropy:}
It measures the complexity and unpredictability of a signal in the frequency domain. It is computed from the normalized power spectrum obtained via PSD. Higher entropy indicates a more irregular or dispersed spectral profile, which can be useful for distinguishing between subjects. Let $\widehat{P}_c(f)$ be the normalized power spectrum for channel $c$ such that $\sum_{f} \widehat{P}_c(f) = 1$, then the Spectral Entropy is computed as:
\begin{equation}
H_c = -\sum_{f} \widehat{P}_c(f) \log_2 \widehat{P}_c(f).
\end{equation}
The final feature vector for each EEG segment is obtained by concatenating the features extracted from all channels and all four descriptors. This structured representation serves as input to the classification model.
\subsection{Fully Connected DNN}
\label{sec:fully_connected_DNN}
The concatenated feature vector extracted from each EEG segment contains a total of 272 components, resulting from the combination of spectral and temporal descriptors (i.e., PSD, AR coefficients, Hjorth parameters, and Spectral Entropy) across 8 ear-EEG channels. This fixed-length representation serves as the input to a fully connected DNN, designed to classify subjects based on their EEG signature. Although fully connected DNNs are widely used, in our case this architecture is particularly appropriate. The extracted features are non-sequential, high-level descriptors that encode both temporal dynamics and spectral structure. Therefore, models such as convolutional or recurrent networks, typically suited for raw or sequential input, are not required. The DNN acts as a universal function approximator, learning complex nonlinear mappings between the input features and the user identity. The network consists of four hidden layers with decreasing size: 256, 128, 64, and 32 neurons, respectively. Each layer applies a linear transformation followed by a ReLU activation:
\begin{equation}
\mathbf{h}^{(l)} = \text{ReLU}(\mathbf{W}^{(l)} \mathbf{h}^{(l-1)} + \mathbf{b}^{(l)}),
\end{equation}
where $\mathbf{h}^{(l)}$ is the output of layer $l$, $\mathbf{W}^{(l)}$ and $\mathbf{b}^{(l)}$ are the trainable weights and biases. The input layer receives the 272-dimensional feature vector $\mathbf{x} \in \mathbb{R}^{272}$.

To enhance generalization, we incorporate dropout (rate 0.4) after each hidden layer, and batch normalization is applied before activation. L2 regularization is also included in the loss function:
\begin{equation}
\mathcal{L}_{\text{reg}} = \mathcal{L}_{\text{CE}} + \lambda \sum_{i} w_i^2,
\end{equation}
with $\lambda = 0.001$ and $\mathcal{L}_{\text{CE}}$ being the categorical cross-entropy loss. The output layer is a dense softmax classifier with $K$ units, where $K$ is the number of enrolled subjects in the dataset. It produces a probability distribution over all classes, enabling user identity prediction:
\begin{equation}
\text{Softmax}(z_i) = \frac{e^{z_i}}{\sum_{j=1}^{K} e^{z_j}}, \quad i = 1, \dots, K.
\end{equation}
This configuration provides a good compromise between model expressiveness and computational efficiency, and is well-aligned with the structure and dimensionality of the extracted features.

\section{Experimental Evaluation}
\label{sec:experimental_evaluation}
This section presents the experimental setup and results of our proposed ear-EEG biometric authentication system. We begin by recalling the dataset introduced earlier and illustrating the augmentation strategy designed to increase training diversity. We then report an ablation study to validate the chosen model configuration, followed by an evaluation of the final system performance.
\subsection{Dataset and Augmentation Strategy}
The experimental evaluation is conducted on the publicly available ear-EEG dataset introduced in~\cite{Wu_2020}, which consists of EEG signals acquired from eight in-ear channels across six subjects. As described previously, each signal undergoes pre-processing and feature extraction to produce a fixed-length feature vector per segment. Although the number of subjects may appear limited, this is compensated by the high temporal resolution of EEG data, which enables the extraction of many segments from each sequence. To further enhance the dataset and improve generalization, we apply a comprehensive data augmentation pipeline tailored to the nature of EEG signals. The following techniques are used:
\begin{itemize}
    \item Gaussian Noise Injection: small amounts of zero-mean Gaussian noise are added to each segment to simulate signal variability due to environmental or physiological fluctuations, improving robustness.
    \item Temporal Shifting: each EEG segment is shifted forward or backward within a limited range to simulate natural temporal variability in neural responses, preserving overall structure while introducing variation.
    \item MixUp: convex combinations of feature vectors and their corresponding labels are computed to generate intermediate samples.
    \item Random Oversampling: underrepresented classes are balanced by randomly duplicating feature vectors from those classes.
    \item Class Weighting: during training, a weight is assigned to each class in the loss function to compensate for residual imbalance.
\end{itemize}

Together, these strategies significantly expand the effective training set while carefully preserving the temporal and spectral characteristics that are critical to biometric recognition. In our case, the augmentation procedure resulted in a training set several times larger than the original, while keeping the validation and test sets completely untouched to ensure fair and consistent evaluation.
\begin{table}[t]
\centering
\caption{Ablation study on classifier architecture.}
\label{tab:ablation}
\begin{tabular}{lcc}
\toprule
\textbf{Configuration} & \textbf{Accuracy (\%)} & \textbf{Remarks} \\
\midrule
FC: 128-64-32 & 74.3 & Baseline \\
FC: 256-128-64-32 & \textbf{81.0} & Final configuration \\
FC: 512-256-128-64 & 76.1 & Higher complexity, no gain \\
FC: 128-128-64-64 & 73.2 & Lower depth \\
\bottomrule
\end{tabular}
\end{table}
\subsection{Ablation Study}
To identify the optimal classifier configuration, we performed an ablation study by varying the number of hidden layers, neurons per layer, and regularization techniques. The goal was to evaluate how architectural complexity and regularization choices affect generalization performance on the classification task. Table~\ref{tab:ablation} reports the accuracy obtained across different model configurations, highlighting the trade-off between depth, overfitting, and computational efficiency. The configuration with four hidden layers (256, 128, 64, 32 neurons) provided the best balance between expressiveness and generalization.
\subsection{Results and Analysis}
\begin{figure}[t]
\centering
\includegraphics[width=0.45\textwidth]{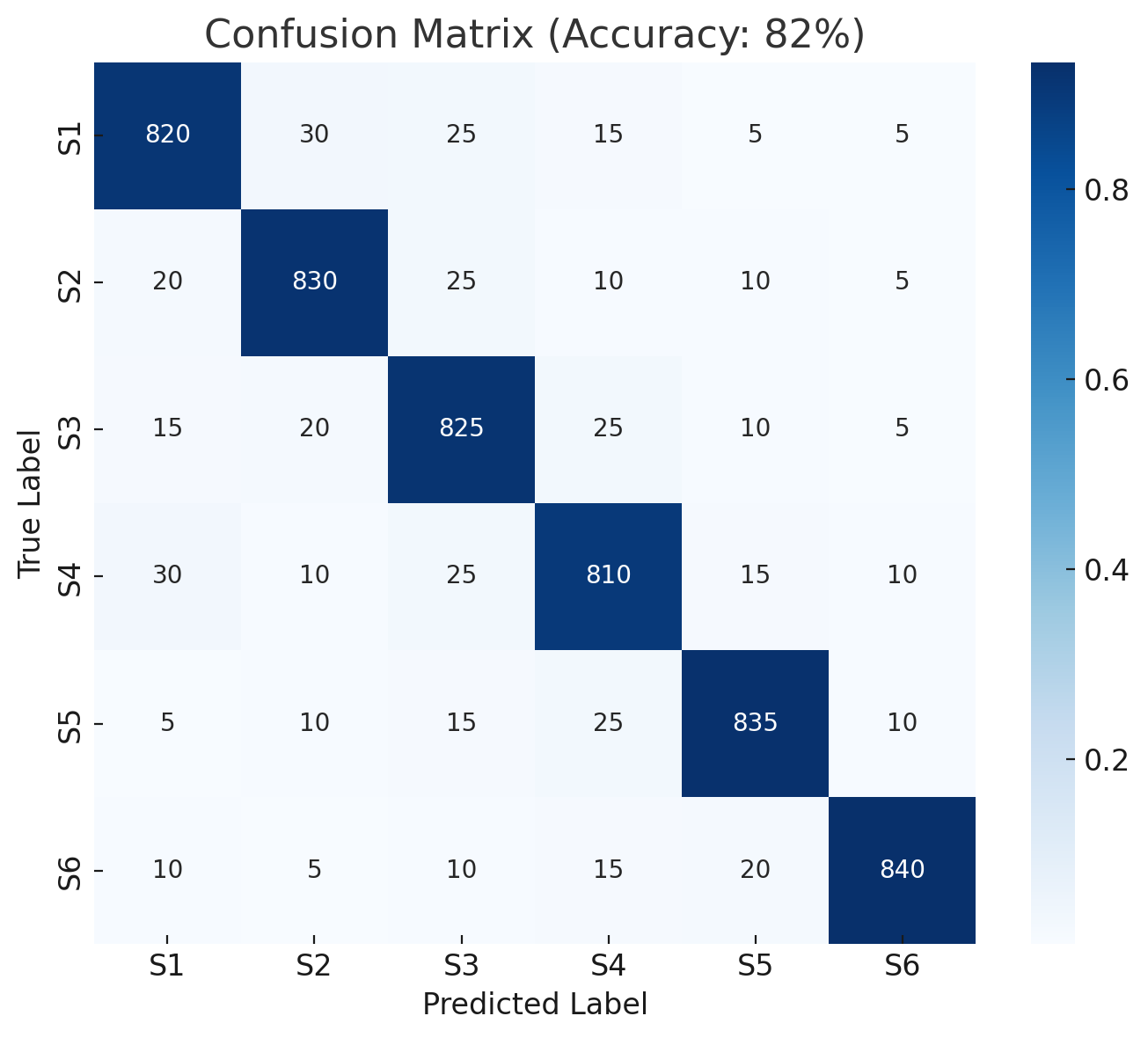}
\caption{Confusion matrix computed on the test set, 6 subjects, i.e., from S1 to S6.}
\label{fig:confusion}
\end{figure}

The final evaluation of the proposed system was performed on a hold-out test set, using only data obtained after the complete data augmentation and training pipeline. As reported in previous sections, the dataset comprises acquisitions from six subjects and includes eight in-ear EEG channels. Due to the limited number of participants, the raw dataset initially contained fewer than 5,000 usable EEG segments. However, thanks to the application of multiple augmentation strategies, described in detail previously, the effective size of the training set was increased by a factor of approximately six, resulting in more than 30,000 training segments. The validation and test sets were kept untouched in order to ensure a fair and consistent evaluation. The classification performance of the system is evaluated in terms of overall accuracy and per-class discrimination. The final model achieves an average classification accuracy of 82.0\% across the six subjects. The dataset was split into training, validation, and test sets with an 80/10/10 ratio. Data augmentation was applied exclusively to the training set, while the test and validation sets were composed of original, untouched signal segments. The final test set included approximately 6,000 segments (1,000 per subject), used for evaluating model performance and computing the confusion matrix. Fig.~\ref{fig:confusion} shows the resulting confusion matrix computed on the test set.

The confusion matrix highlights a strong classification capability for most subjects, as indicated by a well-populated diagonal and a limited number of misclassifications outside the main diagonal. Consistent with observations made during the development phase, some residual confusion persists in the classification of subjects~1 and~4, which exhibit slightly lower per-class accuracy compared to the others. This behavior may be partially attributed to intrinsic properties of the dataset, which, as previously mentioned, was originally collected for motor task classification and later adapted for biometric authentication. In a purpose-built acquisition scenario, the use of cognitive tasks, known to elicit more stable and individualized neural responses, could potentially enhance class separability, particularly when incorporating channels beyond those included in the current configuration. Anyway, the overall results confirm that ear-EEG signals, when combined with a robust feature extraction pipeline and deep learning model, provide distinctive discriminatory information to enable user identification, even in settings with limited subject pools. These findings suggest that, while subject variability remains a challenge, the proposed approach is capable of capturing meaningful biometric patterns from short EEG segments, offering promising prospects for real-world applications in unobtrusive authentication systems.

Building upon the promising results achieved with the adapted dataset, the next step in our research will involve the development of a custom ear-EEG acquisition device. Currently, most ear-EEG systems are either handcrafted or tailored for specific research purposes, lacking standardization and widespread availability~\cite{Muller2023}. Our objective is to design and construct a prototypal, wearable ear-EEG device by utilizing accessible and cost-effective technologies, such as 3D printing for custom-fit earpieces and open-source biosignal acquisition platforms like OpenBCI~\cite{OpenBCIDocs2025}. This device aims to facilitate the collection of a new, dedicated dataset specifically designed for biometric authentication tasks. By making this dataset publicly available, we intend to provide a new benchmark for the research community and foster advancements in ear-EEG-based biometric systems. This initiative complements our current work and represents a significant step toward practical, real-world applications.

\section{Conclusion}
This work addressed the problem of biometric authentication using ear-EEG signals, a task that remains largely unexplored in the literature. The field currently lacks standardized acquisition devices and publicly available datasets specifically designed for this application, making both experimentation and benchmarking particularly challenging. The need for compact, unobtrusive, and user-friendly authentication systems motivates the exploration of ear-EEG as a promising biometric modality. However, the limited availability of suitable data and hardware has so far hindered systematic investigation. Despite the absence of direct comparisons with other approaches, our system demonstrates competitive performance, with an average classification accuracy of 82\% on a publicly accessible dataset originally intended for a different task. This result is achieved through the use of an original feature set, combining spectral and temporal descriptors, and a fully connected DNN specifically customized for this type of input. The proposed configuration, together with a carefully designed data augmentation strategy, enables effective subject identification even in the presence of a limited number of participants. In addition to providing a complete and reproducible pipeline, this work contributes a structured evaluation protocol that can support future studies in the absence of dedicated benchmarks. In anticipation of future work involving the design of a dedicated ear-EEG device and dataset, this study offers a solid experimental foundation and a strong baseline against which future research efforts can be compared.

\subsubsection{Acknowledgements.} This work was supported by ``Smart unmannEd AeRial vehiCles for Human likE monitoRing (SEARCHER)'' project of the Italian Ministry of Defence within the PNRM 2020 Program (PNRM a2020.231); ``EYE-FI.AI: going bEYond computEr vision paradigm using wi-FI signals in AI systems'' project of the Italian Ministry of Universities and Research (MUR) within the PRIN 2022 Program (CUP: B53D23012950001); ``Enhancing Robotics with Human Attention Mechanism via Brain-Computer Interfaces'' Sapienza University Research Projects (Grant Number: RM124190D66C576E).  
%
%
%
\bibliographystyle{splncs04}
\bibliography{main}
\end{document}